\documentclass[conference]{IEEEtran}
\IEEEoverridecommandlockouts
\usepackage{cite}
\usepackage{amsmath,amssymb,amsfonts}
\usepackage{graphicx}
\usepackage{textcomp}
\usepackage{booktabs}
\usepackage{array}
\usepackage{multirow}
\usepackage{xcolor}
\usepackage{url}

\begin{document}

\title{PoseRefer: Pathway-Local Parameters for Semantically Grounded Reference Resolution}

\author{
Anna Deichler\\
KTH Royal Institute of Technology, Stockholm, Sweden\\
deichler@kth.se
}
\maketitle
\begin{abstract} A robot resolving ``put the cup on that one'' must fuse gesture, language, and scene geometry, yet 3D grounding benchmarks only partially capture this regime: descriptions are written post-hoc, gestures are templated, or pointing is staged for the camera. MM-Conv captures natural co-speech gesture from dyadic VR interaction alongside full-body motion capture and 3D scene graphs. We use it to evaluate pose--language fusion with a \emph{decoupled late-fusion} architecture in which pose and text pathways share no learned parameters. The two choices together make category, pose, and text contributions easier to isolate through controlled ablations. Fusion with frozen MiniLM category embeddings exceeds pose alone and the best text-only pathway on every reference type, reaching 31.9\% top-1. The learned scalar gate flips between opposing policies depending on whether the text pathway has category access. This is a reliability diagnostic: fusion-accuracy claims for semantic grounding systems are indistinguishable from category-representation artifacts unless pathways are architecturally decoupled.
\end{abstract}

\section{Introduction}
 
When a person points at a chair and says ``put the cup on that one,'' a service robot observing from a third-person view must integrate gesture, language, and scene geometry to identify the referent. This problem, exocentric reference resolution, has received less attention than its 2D image-grounded cousin, despite being a regime central to many household and assistive robotics scenarios.

Existing 3D grounding benchmarks substitute one or more of the
naturalistic ingredients: post-hoc descriptions on static scans
(ScanRefer~\cite{chen2020scanrefer},
Sr3D/Nr3D~\cite{achlioptas2020referit3d}), template-generated
language with synthetic 2D pose (Ges3ViG~\cite{mane2025ges3vig}),
or single-user pointing performed for the camera
(YouRefIt~\cite{chen2021yourefit}). MM-Conv~\cite{deichler2026mmconv}
differs in kind:  references arose from dyadic VR interaction where two participants worked through a referential task in a shared 3D scene, with full-body motion capture, word-level speech alignment, and ground-truth scene graphs. It contains $\sim\!4{,}000$ references across 5 rooms, with $\sim\!55\%$ accompanied by a detectable pointing or pointing-variant gesture and $\sim\!45\%$ without. Reliable grounding of such references is a prerequisite for any downstream action selection that depends on semantic scene understanding.

\begin{figure}[t]
  \centering
  \includegraphics[width=\columnwidth]{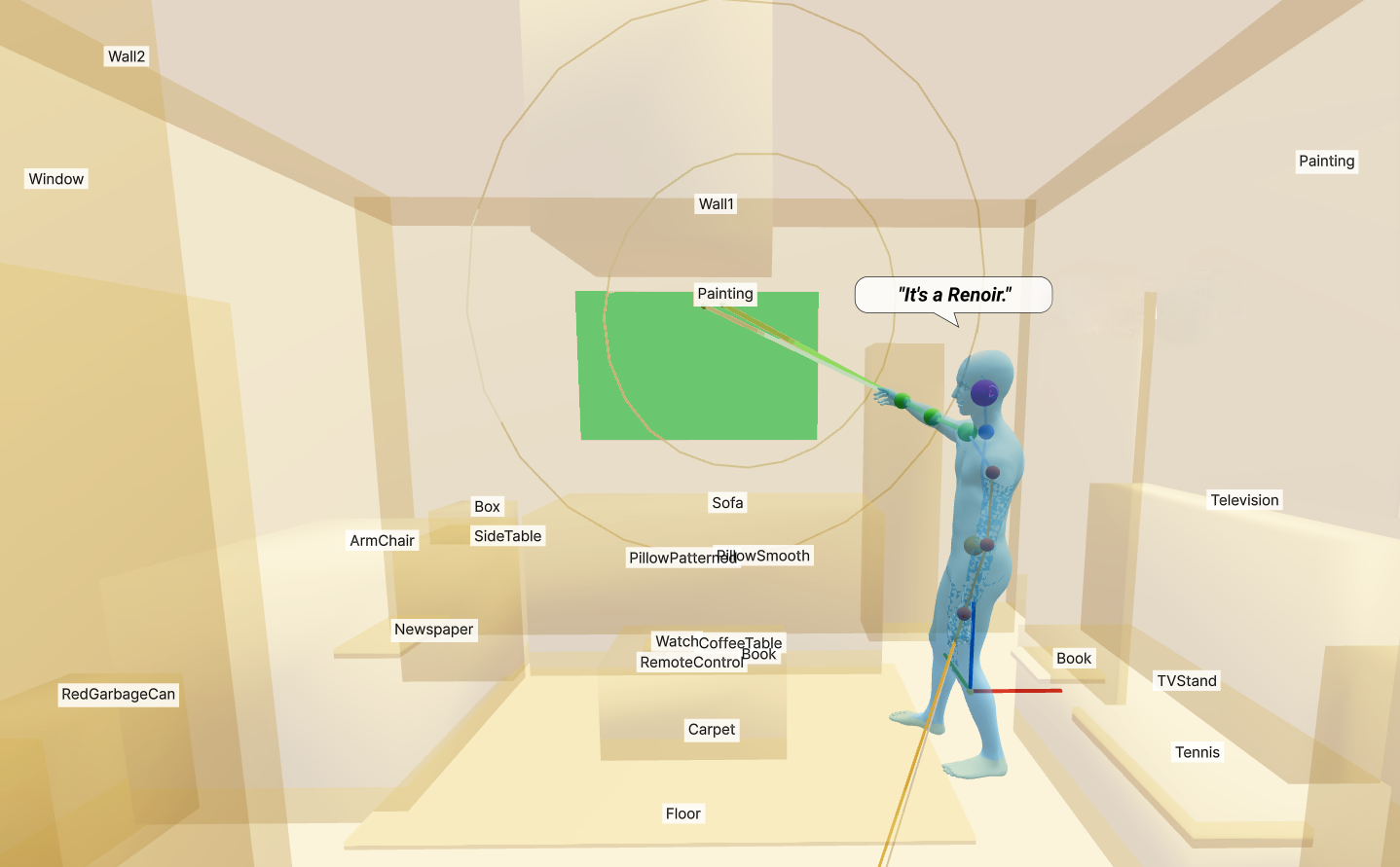}
  \caption{PoseRefer in a 3D scene. The SMPL-X mesh shows the speaker's pose; the pointing ray and head direction define angular proximity to candidate objects (bounding boxes).}
  \label{fig:hero}
\end{figure}

 On naturalistic gesture data, two confounds become visible that synthetic data hides. \emph{First}, per-object category embeddings entangle with all three pathways, pose, text, and the fusion gate, through shared encoders and shared parameter tables. Ablating category in such architectures simultaneously alters every signal and every learned representation, making fusion comparisons hard to interpret. \emph{Second}, the representation chosen for those per-object
categories matters substantially: in our experiments, a learned
16-dimensional category embedding is insufficient at MM-Conv's
per-category training counts, and the fusion gate inherits this noise. Both problems are invisible unless the architecture allows direct ablation of individual signals \emph{and} the category representation itself is varied.

\textbf{Contributions.} 
\begin{enumerate} 
\item \emph{Position:} naturalistic referential gesture data is an important testbed for claims about semantic grounding in multimodal perception. Nearly half of such references occur without clear pointing, a regime synthetic benchmarks largely exclude. 
\item \emph{Method:} a decoupled late-fusion architecture with \emph{pathway-local} category embedding tables (no shared parameters between pose and text), and frozen pretrained semantic embeddings for per-object category on the text pathway. The two choices together reduce category, pose, and text to controllable ablations.
\item \emph{Empirical findings:} on MM-Conv, (a)~the fusion gate commits to opposing policies depending on whether category is available on the text pathway ($\alpha\!=\!0.84 \rightarrow 0.22$), a diagnostic reproducible across cross-validation folds; (b)~pose and text are asymmetrically complementary: pose dominates pointing tiers, text dominates non-pointing tiers; (c)~with pathway-local parameters and frozen MiniLM category embeddings, scalar fusion exceeds both singletons across every reference type, reaching $31.9\%$ aggregate top-1, a gain of 13.1 points over pose alone and 6.9 over the strongest text-only configuration.
\item \emph{Reliability implication:} fusion gains in architectures with shared encoders or shared category embeddings may reflect the category-representation choice rather than properties of fusion itself. Our decoupled setup makes this directly testable and motivates reporting both learned and frozen category representations when evaluating multimodal fusion for semantic grounding.



\end{enumerate}
\section{Related Work}
3D visual grounding operates on post-hoc scene descriptions without the speaker's body (ScanRefer~\cite{chen2020scanrefer}, Sr3D/Nr3D~\cite{achlioptas2020referit3d}, BUTD-DETR~\cite{jain2022bottom}, EDA~\cite{wu2023eda}); Ges3ViG~\cite{mane2025ges3vig} adds template language with synthetic 2D pose and YouRefIt~\cite{chen2021yourefit} captures real but single-user 2D reference. 3D scene graphs have emerged as a compact semantic environment representation for embodied tasks~\cite{hughes2022hydra, gu2024conceptgraphs}; MM-Conv's ground-truth scene graphs let us isolate pose--language fusion from scene-perception error.
Wang et~al.~\cite{wang2024move} use a distance field between skeleton joints and scene points for language-guided motion generation; we adapt the idea to angular distance over discrete candidates. Deichler et al.~\cite{deichler2023pointing} generate
context-aware pointing gestures for embodied agents; we address the inverse problem of interpreting observed gestures for
reference resolution. Related embodied fusion work explores task-adaptive gated routing~\cite{liu2025omnieva} and exocentric reference benchmarks~\cite{islam2025refer360}. Frozen pretrained language spaces appear for semantic matching in open-vocabulary pipelines~\cite{yang2024llm}, but typically as part of a broader vision--language stack rather than an isolated per-object handle in a dedicated fusion pathway.

\section{Dataset and Setup}

MM-Conv  contains approximately $4{,}200$ referring expressions from $6.7$\,h of dyadic VR interaction across 5 AI2-THOR rooms with 42--61 candidate objects per room, synchronized speech (word-level timestamps), full-body SMPL-X parameters at 30\,fps, and 3D scene graphs. Expressions are $38\%$ exact noun phrases, $14\%$ partitives, $48\%$ pronominals (e.g., ``that one'', ``this one''). After filtering for valid targets, text features, and valid temporal windows we evaluate on $3{,}764$ references. We adopt a third-person robot camera view; scene-graph ground truth is unchanged.

\textbf{Category vocabulary.} Scene-graph labels are lowercased, compound labels split, and color/material/size modifiers stripped (``RedVase'' $\rightarrow$ ``vase''), addressing training-signal sparsity in the raw vocabulary. The normalized vocabulary contains $50$ categories. Stripping discards potentially grounding-relevant attributes (e.g., color), a systematic ablation of modifier retention is deferred to future work.

\textbf{Gesture classification.} A random-forest classifier on $16$ SMPL-X kinematic features separates clear pointing from no-gesture (AUC $0.97$). Classifier scores stratify references into four confidence tiers, T1--T4, plus a post-hoc T5 for two-arm pointing; we use the binary partition \emph{pointing} (T1$\cup$T2$\cup$T5, $n\!=\!2{,}068$, $55\%$) vs.\ \emph{non-pointing} (T3$\cup$T4, $n\!=\!1{,}696$) throughout. Tiers are descriptive only, never used as model input. Per-tier counts are reported in Table~\ref{tab:per_tier}.

\section{Method: Decoupled Late Fusion}

\begin{figure}[t]
    \centering
    \includegraphics[width=\linewidth]{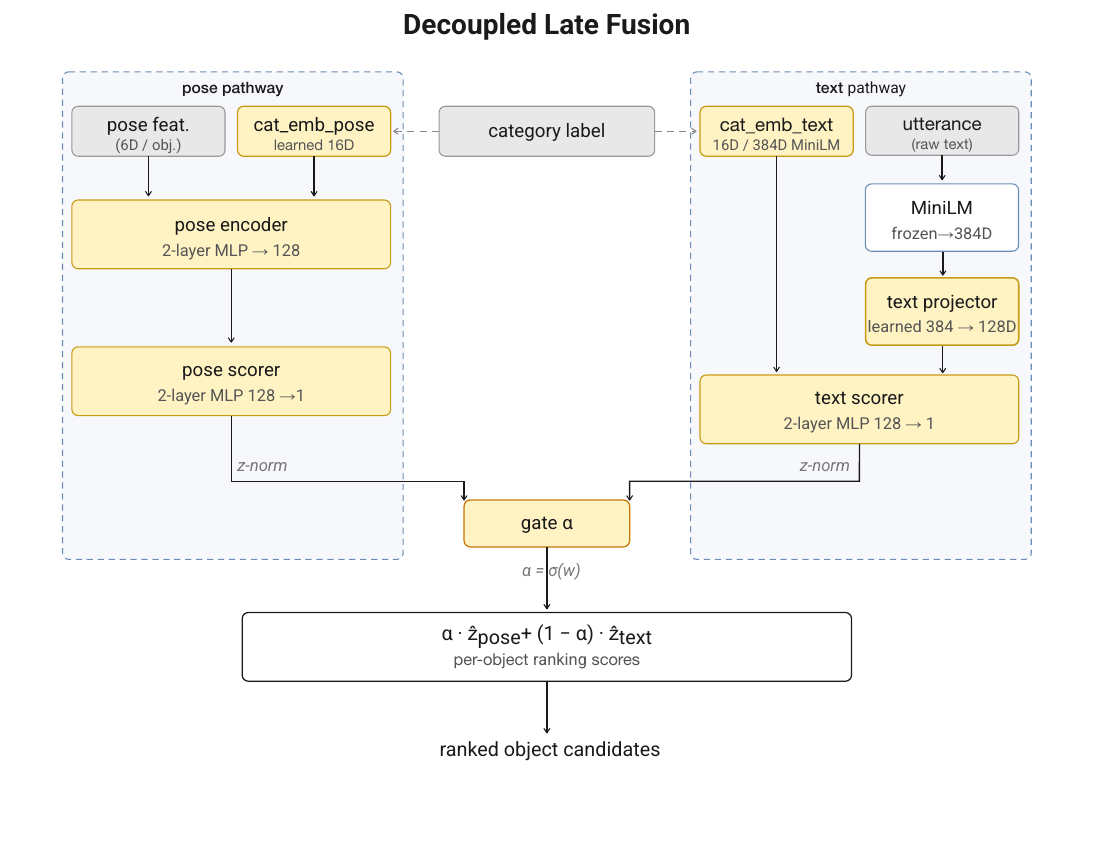}
\caption{PoseRefer architecture. Pose and text pathways operate independently with no shared learned parameters. The text pathway encodes the utterance via a frozen MiniLM-L6 model ($\to$384D) before a learned linear projection to 128D. Each pathway has its own per-object category embedding table: learned 16D on the pose side; on the text side, either learned 16D or (default in PT$_\text{minilm}$) a frozen MiniLM encoding of the canonical category name. Per-pathway scores are $z$-normalized and combined by a learned scalar gate $\alpha$.} 
    \label{fig:architecture}
\end{figure}
We score each candidate object by its angular proximity to the speaker's body channels over a temporal window aligned to the utterance, and, the key architectural choice of this paper, route that score through a pathway that is \emph{architecturally disjoint} from the text pathway and shares no learned parameters with it.

\textbf{Temporal angular affordance.} For object $n$ and channel $j \in \{\text{R arm}, \text{L arm}, \text{head}, \text{body}\}$, the per-frame angle is
$\theta^{(t)}_{n,j} = \angle(\mathbf{d}^{(t)}_j,\mathbf{p}_n - \mathbf{q}^{(t)}_j)$,
where $\mathbf{d}^{(t)}_j$ is the channel direction (pointing ray for arms, facing for head/body), $\mathbf{q}^{(t)}_j$ its origin (wrist, eye midpoint, pelvis), and $\mathbf{p}_n$ the object centroid. Angles
map to scores via a Gaussian kernel $s^{(t)}_{n,j} = \exp(-\theta^2 / 2\sigma_j^2)$ with $\sigma_\text{arm}\!=\!15^\circ$, $\sigma_\text{head}\!=\!30^\circ$,
$\sigma_\text{body}\!=\!45^\circ$.

\textbf{Temporal pooling.} Arm scores are computed on a narrow $\pm\!10$-frame window centered on the utterance hold frame (pointing is near-instantaneous and the hold annotation is accurate); head and body scores use a wider window spanning phrase-start${-}0.5$\,s to phrase-end${+}0.5$\,s (gaze and torso orientation persist through the utterance). We pool with both max and mean for arms (peak vs.\ sustained), max only for head (gaze is spiky), mean only for body (torso is stable), giving a 6-dimensional pose feature vector per object. Arm-extension gating was explored but did not improve
pose-pathway accuracy and is not used in the reported results.

\textbf{Decoupled pathways.} Each candidate object receives scores from two architecturally separate pathways: 
\begin{itemize}
\item \emph{Pose pathway}: the 6-D angular affordance features, 
concatenated with a pose-local $16$-D learned category embedding 
$E_\text{pose}$ (when enabled), are fed to a 2-layer encoder 
($6{+}|E_\text{pose}|\!\rightarrow\!128\!\rightarrow\!128$) followed by  a 2-layer scorer ($128\!\rightarrow\!128\!\rightarrow\!1$) producing a  pose score $s^\text{pose}_n$.
\item \emph{Text pathway}: receives no world-frame geometry, isolating utterance semantics from spatial cues. A frozen MiniLM-L6~\cite{minilm} utterance embedding ($384$-D) is projected to $128$-D, broadcast across objects, concatenated per-object with a text-local category embedding $E_\text{text}$ (when enabled), and passed through a 2-layer scorer ($128{+}|E_\text{text}|\!\rightarrow\!128\!\rightarrow\!1$) producing a text score $s^\text{text}_n$. Per-object, $E_\text{text}$ is either a learned $16$-D vector or a frozen $384$-D MiniLM encoding of the category's canonical name; we ablate both choices.
\end{itemize}
The pathways share no learned parameters: neither MLP weights nor (when learned) category embedding tables.\footnote{Under frozen-MiniLM-on-both, the two pathways read from the same constant buffer, a shared input, not a shared parameter.} Under a shared encoder, or a decoupled encoder with \emph{shared} category embedding, ablating category alters every downstream signal at once and training-time gradients from one pathway shape the embedding seen by the other. Pathway-local embeddings turn ``what does category contribute in each pathway?'' into a controlled experiment.

\textbf{Category representation on the text pathway.} Under a learned $16$-D embedding, each category must learn its representation from LORO-fold (Leave-one-room-out) training examples alone: with $50$ categories and fold training sizes around $3{,}000$, many categories have only a handful of training samples, producing unreliable vectors for the less-frequent labels. A frozen MiniLM encoding of the category name sidesteps this: categories are represented in the same $384$-D semantic space as the utterance embedding, inheriting ``vase''--``chair''--``table'' similarity structure from the pretraining distribution. We treat the choice between learned and frozen MiniLM category embeddings as a methodological ablation; §\ref{ssec:fusion_wins} shows it is also the difference between 28.2\% and 31.9\% aggregate top-1 under fusion.

\textbf{Fusion.} Per-pathway scores are $z$-normalized per sample and
combined with a learned global gate
\[
s_n = \alpha\,\hat{s}^\text{pose}_n + (1-\alpha)\,\hat{s}^\text{text}_n,
\qquad \alpha = \sigma(w),
\]
with $w$ initialized to $0$ so that $\alpha = 0.5$ at the start of training. $\alpha$ is a single learned scalar per training run,
shared across all samples.

\textbf{Training.} LORO cross-validation across the $5$ rooms; $50$ epochs, AdamW, lr $10^{-3}$, batch $32$, cosine schedule; cross-entropy over candidates within each sample. All numbers are reported as mean$\pm$std across three random seeds under LORO cross-validation; per-fold $\alpha$ values are listed where relevant. Dropout 0.3 in all MLPs; weight decay $10^{-4}$. Paired \emph{t}-tests ($n=3$) compare seed-averaged LORO fold means.


\section{Empirical Findings}

Table~\ref{tab:main} reports aggregate accuracy for eight configurations spanning combinations of pathway inclusion and category representation; Tables~\ref{tab:per_tier} and~\ref{tab:per_type} stratify the headline configurations by gesture tier and reference type.

\begin{table}[t]
\caption{Reference resolution accuracy under leave-one-room-out cross-validation. Mean$\pm$std over three random seeds.}
\label{tab:main}
\centering
\small
\setlength{\tabcolsep}{3pt}
\newcommand{\ms}[2]{#1{\tiny$\pm$#2}}
\begin{tabular}{lcccrrr}
\toprule
Config & Pose & Text & Cat & Top-1 & Top-5 & $\alpha$ \\
\midrule
Random                  & --  & --  & --        & $\sim$2.0 & $\sim$10.0 & -- \\
\midrule
P                       & \checkmark & --         & --        & \ms{18.8}{0.3} & \ms{51.7}{0.3} & -- \\
P$_\text{cat}$          & \checkmark & --         & L pose   & \ms{18.9}{0.6} & \ms{49.3}{0.9} & -- \\
T                       & --         & \checkmark & L text   & \ms{22.5}{2.4} & \ms{49.2}{0.4} & -- \\
T$_\text{minilm}$       & --         & \checkmark & M text   & \ms{25.0}{0.4} & \ms{50.5}{0.4} & -- \\
PT$_\text{nocat}$       & \checkmark & \checkmark & --       & \ms{18.9}{0.1} & \ms{51.7}{0.2} & \ms{0.837}{0.000} \\
PT                      & \checkmark & \checkmark & L both   & \ms{28.2}{0.2} & \ms{57.8}{0.9} & \ms{0.222}{0.001} \\
\textbf{PT$_\text{minilm}$} & \checkmark & \checkmark & M text & \textbf{\ms{31.9}{0.1}} & \textbf{\ms{60.2}{0.5}} & \ms{0.248}{0.007} \\
PT$_\text{minilm,both}$ & \checkmark & \checkmark & M both   & \ms{31.6}{0.1} & \ms{59.7}{0.4} & \ms{0.267}{0.002} \\
\bottomrule
\end{tabular}
\end{table}

Three findings structure the analysis. First (\S\ref{ssec:flip}), the learned fusion gate flips between opposing policies depending solely on whether the text pathway has a per-object category handle, a diagnostic that the pathway-local decoupling makes directly observable. Second (\S\ref{ssec:complement}), pose and text are strongly asymmetrically complementary across gesture tiers, establishing the disagreement pattern fusion must exploit. Third (\S\ref{ssec:fusion_wins}), with frozen MiniLM category embeddings on the text pathway, scalar fusion exceeds both singletons across every reference type, reaching $31.9\%$ top-1.

\subsection{The Fusion Gate Flips with Category Presence} \label{ssec:flip}

The learned global gate commits to opposing policies depending on whether the text pathway has access to a per-object category handle. Under PT$_\text{nocat}$ (no category on either pathway), the gate settles at $\alpha\!=\!0.837$ across folds, trusting pose heavily. Under PT (learned category embeddings on both pathways), the same architecture on the same $3{,}764$ references under the same LORO folds settles at $\alpha\!=\!0.222$, nearly the opposite policy. Only the text-pathway category handle differs between the two runs; every other hyperparameter, random seed, and fold assignment is held fixed.

The gate does not ``drift'' to its final $\alpha$; its final value is highly reproducible across the five LORO folds (per-fold $\alpha$: PT$_\text{nocat}$ in $[0.831, 0.842]$; PT in $[0.194, 0.237]$). This is a deterministic architectural response, not a training-noise effect. The interpretation is mechanical: when the text pathway has no per-object distinguishing signal beyond a broadcast utterance embedding, architecturally the case here, since the text scorer's input $\text{text\_projector}(\text{utterance})$ is identical across all candidate objects within a sample, the gate correctly discounts it and trusts pose. When category makes the text pathway per-object-informative, the gate correctly shifts its weight to text.

This has two consequences. Scalar fusion gates are sensitive to architectural choices that determine whether each pathway carries per-object signal: small representation changes produce qualitative changes in gate policy. And a single global $\alpha$ commits to one tradeoff across the full dataset, which (as \S\ref{ssec:complement} will show) is optimal for neither the pointing nor the non-pointing regime. Both observations frame the analysis that follows.

\subsection{Asymmetric Pose--Text Complementarity}
\label{ssec:complement}



\begin{table}[t]
\centering
\caption{Top-1 accuracy by gesture tier. Mean$\pm$std over three random seeds under LORO cross-validation.}
\label{tab:per_tier}
\begin{tabular}{lccccc}
\toprule
Config & T1 & T2 & T3 & T4 & T5 \\
$n$ & 1{,}344 & 558 & 153 & 1{,}543 & 166 \\
\midrule
P                       & 31.5$\pm$0.6 & 20.4$\pm$0.7 & \phantom{0}9.6$\pm$0.8 & \phantom{0}7.5$\pm$0.3 & 24.3$\pm$0.3 \\
T\textsubscript{minilm} & 22.8$\pm$0.5 & 25.0$\pm$0.4 & 34.0$\pm$1.4 & 26.5$\pm$0.4 & 20.9$\pm$1.6 \\
PT\textsubscript{minilm}& 37.1$\pm$0.9 & 35.7$\pm$0.6 & 33.3$\pm$1.4 & 25.4$\pm$0.3 & 35.9$\pm$1.6 \\
\bottomrule
\end{tabular}

\vspace{0.3em}
\footnotesize
Notes: T1/T2/T5 contain stronger pointing cues; T3/T4 are weaker or non-pointing tiers.
\end{table}

Pose and text disagree systematically by gesture tier (Table~\ref{tab:per_tier}). On pointing references (T1$\cup$T2$\cup$T5, $n\!=\!2{,}068$), pose alone achieves $27.9\%$ top-1; text (T$_\text{minilm}$) trails at $23.2\%$. On non-pointing references (T3$\cup$T4, $n\!=\!1{,}696$), the pattern reverses: pose collapses to $7.7\%$ while text reaches $27.2\%$. The per-tier view sharpens the asymmetry: on T1 (clear pointing), pose leads text by $8.7$ points ($31.5$ vs.\ $22.8$); on T4 (no gesture), text leads pose by $19$ points ($26.5$ vs.\ $7.5$). Pose stays above random on T4 ($7.5\%$ vs.\ $\sim\!2\%$) because the head and body channels pick up non-deictic orientation cues even when arms are at rest. Neither singleton exceeds $27\%$ on the full dataset.

An \emph{oracle} policy that selects the best singleton per subset (pose on pointing, text on non-pointing) achieves a weighted top-1 of $\frac{2068}{3764}\cdot 27.9 + \frac{1696}{3764}\cdot 27.2 \approx 27.6\%$. Any fusion method that does not exceed this oracle is doing nothing beyond implicit subset selection.

\subsection{Scalar Fusion Exceeds Singletons When Pathways Are Well-Specified}
\label{ssec:fusion_wins}

\begin{table}[t]
\centering
\caption{Top-1 accuracy by reference type. Mean$\pm$std over three random seeds under LORO cross-validation.}
\label{tab:per_type}
\begin{tabular}{lccc}
\toprule
Config & Exact NP & Pronom. & Part. \\
$n$ & 1{,}433 & 1{,}810 & 521 \\
\midrule
P                       & 19.2$\pm$0.1 & 18.2$\pm$0.5 & 19.9$\pm$0.7 \\
T\textsubscript{minilm} & 37.4$\pm$0.8 & 17.3$\pm$0.3 & 17.9$\pm$0.5 \\
PT\textsubscript{minilm}& 45.2$\pm$0.2 & 23.3$\pm$0.1 & 25.3$\pm$0.9 \\
\bottomrule
\end{tabular}

\vspace{0.3em}
\footnotesize
Notes: Exact NP = exact noun phrase; Pronom. = pronominal reference; Part. = partitive reference.
\end{table}

PT$_\text{minilm}$, fusion with frozen MiniLM category embeddings on the text pathway and learned $16$-D on the pose pathway, reaches $31.9\%$ top-1 aggregate, exceeding pose ($18.8\%$) by $13.1$ points and T$_\text{minilm}$ ($25.0\%$) by $6.9$ (both significant, paired $t$-tests $p<0.01$). It also exceeds the singleton-oracle ($27.6\%$) by $4.3$ points: the gate is combining complementary scores within each sample in a way that simple subset-switching cannot reproduce.

PT$_\text{minilm}$ wins on every reference type (Table~\ref{tab:per_type}). The margin is largest on exact noun phrases, where the utterance contains a matching scene-graph label: $45.2\%$, exceeding T$_\text{minilm}$ by $7.8$ points. On pronominals ($23.3\%$) and partitives ($25.3\%$), which carry no explicit category information, PT$_\text{minilm}$ still exceeds both singletons, indicating that the benefit is not purely lexical matching. The fusion gain is visible even near pose's geometric ceiling: on T1 (clear pointing), pose alone reaches $31.5\%$ and PT$_\text{minilm}$ lifts this to $37.1\%$, text and pose jointly disambiguating among the pointed-at candidates in the pointing cone.

\textbf{Category representation, not just category presence, shapes fusion behavior.} PT with learned $16$-D category embeddings reaches $28.2\%$ aggregate, $3.7$ points below PT$_\text{minilm}$  (paired $t(2)=12.4$, $p=0.005$) despite identical architecture and training. Yet on text alone, learned (T, $22.5\%$) and MiniLM (T$_\text{minilm}$, $25.0\%$) are indistinguishable ($p=0.323$), note the substantially higher seed variance under learned embeddings, consistent with unreliable per-category vectors at MM-Conv's training counts. The representation-choice gap \emph{emerges specifically under fusion}, where the utterance embedding and per-object category both reside in MiniLM space for PT$_\text{minilm}$, letting the gate combine pathways more sharply. PT$_\text{minilm,both}$ (MiniLM on both pathways) reaches $31.6\%$, within noise of PT$_\text{minilm}$: the pose pathway's $6$-D angular features do not benefit from a $384$-D semantic augmentation.


\section{Implications for Reliable Semantic Grounding}

\textbf{Category representation is a hidden confound in fusion evaluation.} The choice of per-object category representation (learned $16$-D vs.\ frozen MiniLM $384$-D) changes fusion accuracy by $3.7$ points while leaving text-alone accuracy unchanged, a dissociation only observable when category contribution is isolated per pathway. Under shared-encoder or shared-category-table architectures this effect is indistinguishable from a fusion-mechanism limitation. For semantic grounding systems whose downstream action selection depends on fusion reliability, both category representations should be reported, and pathway-local decoupling treated as a prerequisite for interpretable fusion claims.

\textbf{Naturalistic referential gesture data is the right testbed.} Synthetic and post-hoc benchmarks exclude the $\sim\!45\%$ of references without clear gestures, eliminating the complementarity signal fusion can exploit. Fusion methods evaluated only on synthetic or post-hoc data may behave differently under naturalistic communicative pressure, where roughly half of references lack clear pointing.

\textbf{Limitations and future work.} Our analysis is bounded by the use of ground-truth SMPL-X; performance under estimated pose is the logical next step for deployment. While scalar fusion significantly exceeds singletons, the global gate is suboptimal for non-pointing tiers; per-sample gating conditioned on pointing-classifier output would allow $\alpha \!\to\! 0$ on non-deictic expressions. Comparisons against open-vocabulary VLM-based 3D grounding would further contextualize absolute accuracy.
\bibliographystyle{IEEEtran}
\bibliography{ref}
\end{document}